\documentclass[sigconf]{acmart}
\AtBeginDocument{%
  }

\acmISBN{978-1-4503-XXXX-X/2018/06}

\usepackage{booktabs}
\usepackage{algorithm}
\usepackage{algorithmic}
\usepackage{natbib} 
\usepackage{caption} 
\usepackage{multirow}
\usepackage{amsmath}
\usepackage{makecell}
\usepackage{booktabs}
\usepackage{xcolor}  % 引入颜色支持
\usepackage{amssymb}  % 提供箭头符号支持
\usepackage{longtable}
\usepackage{epstopdf}
\usepackage{graphicx}
\usepackage{times}  % DO NOT CHANGE THIS
\usepackage{helvet}  % DO NOT CHANGE THIS
\usepackage{courier}  % DO NOT CHANGE THIS
\usepackage{enumitem}

\begin{document}
\title{Chain or tree? Re-evaluating complex reasoning from the perspective of a matrix of thought.}

% \author{
%     Fengxiao Tang\textsuperscript{\rm 1},
%     Yufeng Li\textsuperscript{\rm 1},
%     Zongzong Wu\textsuperscript{\rm 1},
%     Ming Zhao\textsuperscript{\rm 1}
% }
% \affiliations{
%     \textsuperscript{\rm 1}School of Computer,Science and Engineering,Central South University, Changsha, China\\
%     tangfengxiao@csu.edu.cn, 107552404871@stu.xju.edu.cn, Wzy\_Yeah@csu.edu.cn, meanzhao@csu.edu.cn
% }

% \author{Fengxiao Tang}
% \authornote{Both authors contributed equally to this research.}
% \email{trovato@corporation.com}
% \orcid{1234-5678-9012}
% \author{G.K.M. Tobin}
% \authornotemark[1]
% \email{webmaster@marysville-ohio.com}
% \affiliation{%
%   \institution{Institute for Clarity in Documentation}
%   \city{Dublin}
%   \state{Ohio}
%   \country{USA}
% }

\author{Fengxiao Tang}
\affiliation{%
  \institution{School of Computer,Science and Engineering,Central South University}
  \city{Changsha}
  \country{China}}
\email{tangfengxiao@csu.edu.cn}

\author{Yufeng Li}
\affiliation{%
  \institution{School of Computer,Science and Engineering,Central South University}
  \city{Changsha}
  \country{China}}
\email{107552404871@stu.xju.edu.cn}

\author{Zongzong Wu}
\affiliation{%
  \institution{School of Computer,Science and Engineering,Central South University}
  \city{Changsha}
  \country{China}}
\email{Wzy\_Yeah@csu.edu.cn}

\author{Ming Zhao}
\affiliation{%
  \institution{School of Computer,Science and Engineering,Central South University}
  \city{Changsha}
  \country{China}}
\email{meanzhao@csu.edu.cn}

% \author{Valerie B\'eranger}
% \affiliation{%
%   \institution{Inria Paris-Rocquencourt}
%   \city{Rocquencourt}
%   \country{France}
% }

% \author{Aparna Patel}
% \affiliation{%
%  \institution{Rajiv Gandhi University}
%  \city{Doimukh}
%  \state{Arunachal Pradesh}
%  \country{India}}

% \author{Huifen Chan}
% \affiliation{%
%   \institution{Tsinghua University}
%   \city{Haidian Qu}
%   \state{Beijing Shi}
%   \country{China}}

% \author{Charles Palmer}
% \affiliation{%
%   \institution{Palmer Research Laboratories}
%   \city{San Antonio}
%   \state{Texas}
%   \country{USA}}
% \email{cpalmer@prl.com}

% \author{John Smith}
% \affiliation{%
%   \institution{The Th{\o}rv{\"a}ld Group}
%   \city{Hekla}
%   \country{Iceland}}
% \email{jsmith@affiliation.org}

% \author{Julius P. Kumquat}
% \affiliation{%
%   \institution{The Kumquat Consortium}
%   \city{New York}
%   \country{USA}}
% \email{jpkumquat@consortium.net}

%%
%% By default, the full list of authors will be used in the page
%% headers. Often, this list is too long, and will overlap
%% other information printed in the page headers. This command allows
%% the author to define a more concise list
%% of authors' names for this purpose.
\renewcommand{\shortauthors}{Trovato et al.}

%%
%% The abstract is a short summary of the work to be presented in the
%% article.
\begin{abstract}
Large Language Models (LLMs) face significant accuracy degradation due to insufficient reasoning ability when dealing with complex and abstract tasks. Thought structures such as Chain of Thought (CoT) and Tree of Thought (ToT) focus on enhancing the reasoning capability of LLMs. However, they suffer from inherent drawbacks such as redundancy within the same layer of the tree structure and the singularity of the paths in the chain structure. Some studies have utilized Retrieval-Augmented Generation (RAG) methods to enhance CoT and ToT in mitigating hallucinations in LLMs, yet the fundamental shortcomings of the thought structures still persist. Furthermore, when dealing with multi-entity and multi-hop information, the retrieved verification knowledge often contains large amounts of fragmented, superficial, or even erroneous data, misleading the reasoning process of LLMs. To address these issues, we propose the Matrix of Thought (MoT), a novel and efficient thought structure for LLMs. MoT explores problems in both horizontal and vertical dimensions through a "column-cell communication" mechanism, enabling LLMs to actively engage in multi-strategy and deep thinking while reducing redundancy in the thought nodes within the column cells, thereby enhancing the reasoning capability of LLMs. Additionally, through a fact-correction mechanism, it leverages the knowledge graph triples retrieved by RAG and the original text to construct knowledge units and correct erroneous answers. To validate the effectiveness of this method, we conducted extensive experiments in three tasks: 24-point game, question answering evaluation, and proposition writing. The results demonstrate that our framework outperforms state-of-the-art methods, with reasoning time only 14.4\% of that of the baseline method, proving its efficiency and accuracy.
\end{abstract}
%% A "teaser" image appears between the author and affiliation
%% information and the body of the document, and typically spans the
%% page.

\received{20 February 2007}
\received[revised]{12 March 2009}
\received[accepted]{5 June 2009}

%%
%% This command processes the author and affiliation and title
%% information and builds the first part of the formatted document.
\maketitle

\section{Introduction}
The process by which humans move from learning to practice often involves understanding and reasoning through various pieces of information, leading to decisions that align with or even go beyond the content described in the original information. Recent studies have shown that Large Language Models (LLMs) can achieve a similar process of learning to practice in complex fields such as intelligence analysis\cite{wu2024kgv} and creative writing\cite{qin2024charactermeet}. However, because they handle problems and generate thoughts through token-level predictions, this approach limits their ability to expand high-level or multi-perspective reasoning\cite{huang2022inner}.

% The human transition from learning to practice generally entails comprehending and reasoning over diverse information sources, followed by making decisions that align with—or even extend beyond—the original information. Recent studies have demonstrated that large language models (LLMs) can replicate this learning-to-practice paradigm in complex domains such as intelligence analysis \cite{wu2024kgv} and creative writing \cite{qin2024charactermeet}. Complex question answering (QA) is a foundational and critical downstream task for LLMs \cite{zhuang2023toolqa}; it demands that a QA system command extensive knowledge and perform sophisticated reasoning over that knowledge to deliver satisfactory answers.

\begin{figure*}[t!]
\centering
\includegraphics[width=2\columnwidth]{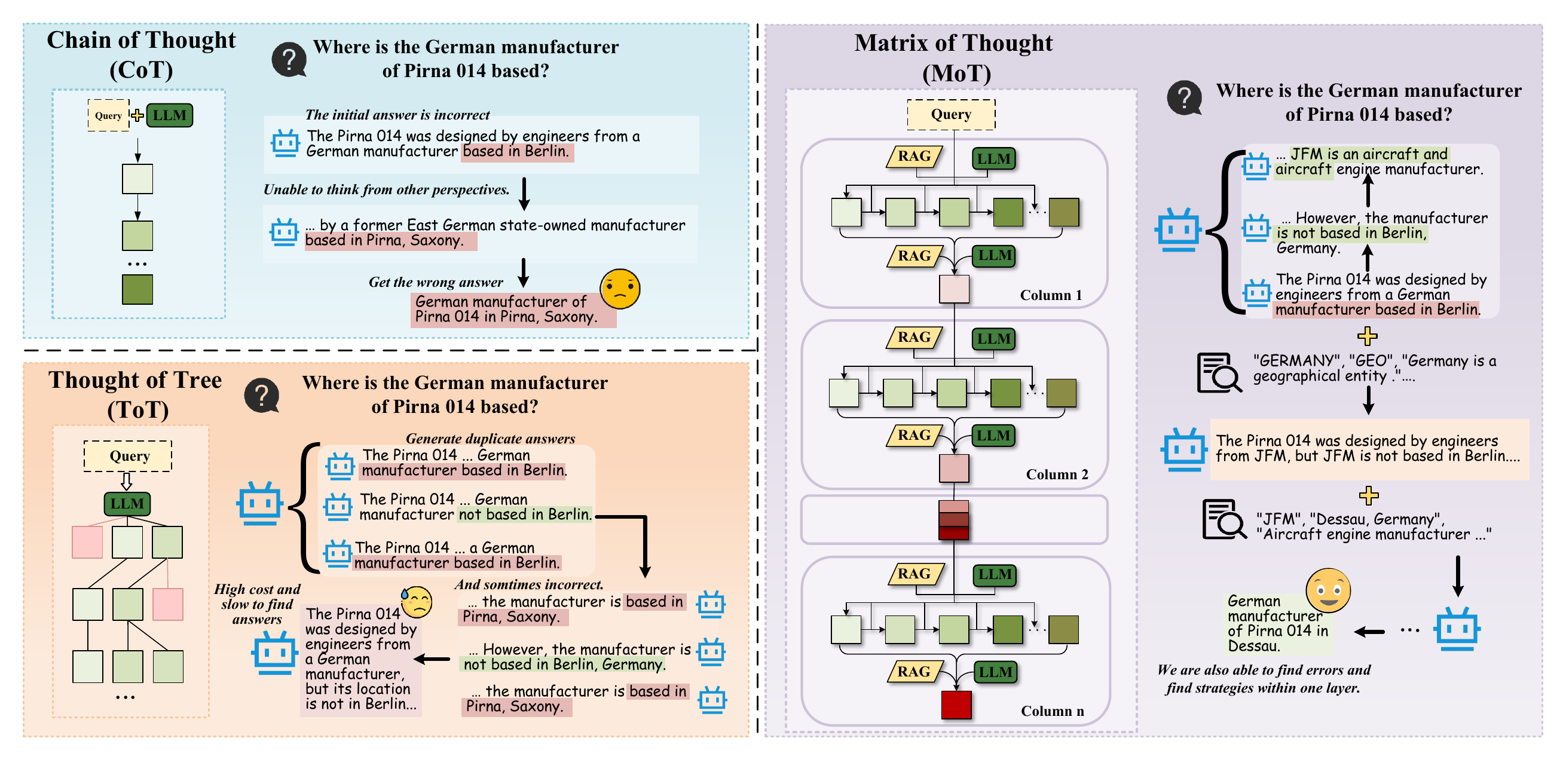}
\vspace{-0.5cm}
\caption{Comparison of Chain of Thought, Thought of Tree, and MoT for Complex Question Answering.}
\label{fig:MOT_V2}
\end{figure*}

Some studies attempt to enhance the reasoning performance of LLMs through thought structures. For example, Chain-of-Thought (CoT) \cite{wei2022chain}, as shown in Figure \ref{fig:MOT_V2}, decomposes the reasoning process of LLMs into a series of ordered reasoning steps, thereby enhancing its reasoning capability. However, its reasoning path is singular, unable to explore multiple paths within the solution space, and errors in reasoning may not be corrected in time, leading to deviations. Tree-of-Thought (ToT) \cite{yao2023tree} structures the reasoning process into a hierarchical tree model, allowing the model to reason across multiple levels and branches. However, its node selection mechanism is not efficient, and it generates a large number of redundant nodes, resulting in resource waste and inefficient reasoning.

% Recently, large language models (LLMs) have become one of the most promising and efficient solutions in the QA field, thanks to their massive pre-trained knowledge and the reasoning abilities they autonomously generate during training\cite{guo2025deepseek}. However, some studies indicate that LLMs exhibit significant shortcomings in complex question-answering reasoning tasks. Various research efforts have attempted to enhance LLMs' reasoning performance using thought structures. For example, the Chain-of-Thought (CoT) framework\cite{wei2022chain} breaks down the reasoning process of LLMs into a series of ordered steps, thereby improving their reasoning capabilities. However, CoT suffers from a single reasoning path, limiting its ability to explore multiple reasoning paths within the solution space, and may also result in incorrect ideas that cannot be corrected in time, leading to deviations. Tree-of-Thought (ToT)\cite{yao2023tree} organizes the reasoning process into a hierarchical tree structure, allowing the model to reason across multiple layers and branches. However, its node selection mechanism is inefficient and often produces redundant nodes, leading to resource wastage and ineffective reasoning.

As the complexity of tasks increases, methods such as CoT and ToT are prone to deviations from the facts during the reasoning process due to hallucinations [citation], which cannot be corrected in time, leading to a misalignment in the subsequent reasoning process. Some works employ Retrieval-Augmented Generation (RAG) to guide and correct the responses of LLMs by retrieving external knowledge to supplement the models, followed by generation to produce more accurate and reliable answers\cite{cuconasu2024power,fan2024survey}.Currently, mainstream reasoning methods embed RAG into thought structures to provide LLMs with the ability to verify facts. RAT\cite{wang2024rat} breaks down the stepwise reasoning process of CoT and adds the RAG correction process to each intermediate reasoning step to ensure that the reasoning trajectory does not deviate. RATT\cite{zhang2025ratt} combines the multi-branch, multi-level reasoning of the tree structure with RAG’s external knowledge to correct factual errors, thereby enhancing the reasoning ability of LLMs. However, in the aforementioned methods, RAG simply supports knowledge based on similarity matching, and the retrieved verification knowledge often lacks coherence, creating fragmentation and misalignment with the LLM's preferences. Additionally, when performing complex question-answering tasks, especially open-domain tasks involving multiple entities and multi-hop information, the retrieved verification knowledge often contains large amounts of irrelevant or even erroneous information, thus misleading the reasoning process of LLMs.

To achieve the aforementioned goal, we propose a novel Matrix of Thought (MoT) method, which ensures that the reasoning process can proceed with multiple branches and low redundancy in the context of solving complex problems. As shown in Figure \ref{fig:MOT_V2}, in MoT, we use RAG to retrieve KG triples and their corresponding original text segments that match the question, which are then integrated as auxiliary knowledge for initiating reasoning and fact verification. Additionally, we introduce a column communication mechanism that actively helps the LLM explore other perspectives for solving the problem, thereby reducing redundancy in the generated reasoning process within the column. Subsequently, MoT combines the various branch strategies with the auxiliary knowledge from RAG, summarizes and optimizes them, and uses this as the reasoning initiation knowledge for the next column unit, thereby converging towards the most promising answer.

% To achieve these goals, we propose a novel Matrix of Thought (MoT) approach to ensure that the reasoning process for complex question-answering tasks is multi-branching and low-redundancy. As shown in Figure \ref{fig:MOT_V2}, in MoT, our RAG retrieves entity-relation-entity triples that match the question, along with their corresponding original text segments, which are then constructed into knowledge units to serve as auxiliary knowledge for initiating reasoning and fact verification. Additionally, we introduce a column cell communication mechanism that actively helps the LLM explore alternative perspectives for solving the problem, thereby reducing the redundancy in the generated reasoning process within the column. Then, MoT combines the various branching strategies with RAG's auxiliary knowledge for summarization and optimization, using this as the reasoning initiation knowledge for the next column cell, thereby approaching the most promising answer.

The contributions of this paper are summarized as follows:
\begin{itemize}[leftmargin=*, itemsep=0pt, parsep=0pt, topsep=0pt]
\item  We propose an innovative thought structure paradigm, MoT, for large model reasoning. It addresses the issues of information redundancy within the same layer of tree structures and the singularity of decision paths in chain structures during reasoning, enhancing the flexibility and adaptability of reasoning with minimal cost.

\item We explore a novel method of embedding RAG within the LLM thought structure, constructing knowledge units to replace the conventional naive knowledge in RAG, in order to align with the LLM's understanding preferences and enhance the integrity of the retrieved knowledge.

\item We conducted a series of experiments to validate the efficiency and reliability of our method in various complex tasks. The results demonstrate that our method holds a clear advantage over existing approaches.
\end{itemize}

\section{Related Work}
\subsection{Thought Structures for LLMs}
The thought structures proposed \cite{ding2023everything,minaee2024large} is a mainstream prompt engineering approach that guides LLMs to engage in more logical thought processes, thereby improving the quality of generated responses. Chain of Thought \cite{wei2022chain} is a representative work that guides LLMs to generate intermediate reasoning steps within the prompt, thereby steering the large model through the reasoning process. Wang et al. developed the Self-consistency with Chain of Thought (CoT-SC) method \cite{wang2023scott}, which extends CoT by solving a problem using multiple independent and distinct chains of thought. The most reliable answer is then selected as the final output response. The Tree of Thoughts (ToT) \cite{yao2023tree} abstracts the reasoning process of a problem into a tree search. By constructing multiple branches and evaluating their contributions to solving the problem, ToT guides the LLM in reasoning, enabling a broad and global exploration of the solution space. 

Current mainstream research focuses on CoT and ToT; however, the inherent properties of their structures present performance bottlenecks, leading to issues such as single reasoning paths, generation of redundant information, and the risk of deviating from the reasoning trajectory due to incorrect ideas during inference. One of the focuses of our research is to develop a more efficient and reliable thought structure paradigm, enabling the LLM to achieve superior performance in various complex tasks within this paradigm.

\subsection{Retrieval-Augmented Generation (RAG) for LLM}
When LLMs solve extremely complex tasks, they suffer from "complete accuracy collapse"\cite{illusion-of-thinking}, significant performance degradation, and LLMs hallucination problems due to insufficient reasoning capabilities\cite{sriramanan2024llm}. Retrieval-Augmented Generation (RAG)\cite{lewis2020retrieval} is a pivotal technique for mitigating such hallucinations and enhancing the quality of LLM outputs. Retrieval-Augmented Generation (RAG) stores task-relevant knowledge in an external repository and, when the LLM is invoked, retrieves pertinent information to serve as supplementary context. GraphRAG \cite{edge2024local} extends this paradigm by incorporating knowledge graphs (KGs): an LLM produces concise community-level summaries that both abstract the input documents and provide graph indices; each community then generates its response independently, and the individual outputs are subsequently aggregated into a global answer. Recent studies have begun to integrate explicit reasoning paradigms into the RAG workflow. Retrieval-Augmented Thought (RAT)\cite{wang2024rat} merges the RAG retrieval step directly into the chain-of-thought process, allowing the model to iteratively refine and correct its reasoning step by step. Retrieval-Augmented Thought Tree (RATT) \cite{zhang2025ratt} adopts a tree-structured reasoning framework in which each node invokes RAG for factual verification, thereby enhancing the logical soundness and factual reliability of LLM reasoning.

However, current research often treats the retrieved knowledge as various fragmented KG triples or text, which do not align well with the preferences of LLMs. Additionally, there has been limited work on designing RAG systems specifically for assisting thought structure reasoning. This paper focuses on studying the optimal representation of the knowledge retrieved by RAG to aid in reasoning within thought structures, enabling LLMs to efficiently utilize the knowledge for reasoning and derive the best possible answers.

\section{Methodology}
\begin{figure*}[h]
\centering
%\vspace{-1.2cm}
\includegraphics[width=2\columnwidth]{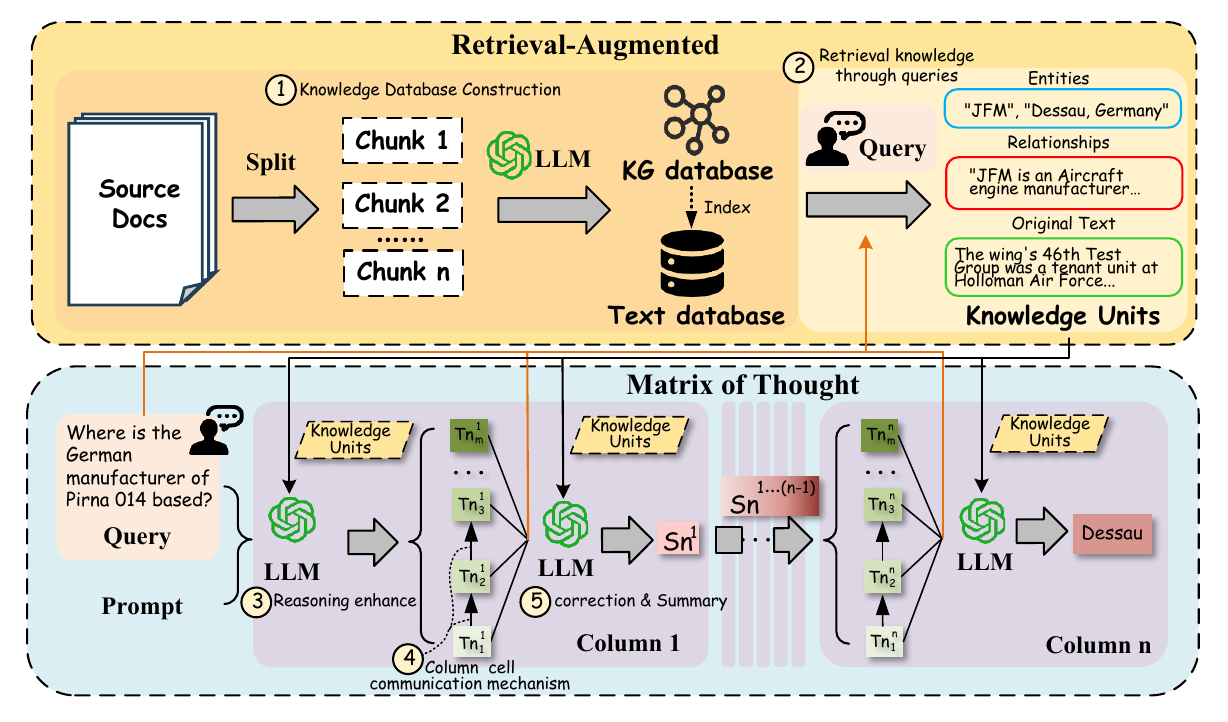}
\caption{Framework of MoT. Given an input including a query and source documents, the LLM retrieves relevant knowledge units from both knowledge graphs and text databases. These knowledge units are then integrated into the Matrix of Thought (MoT), where multiple reasoning strategies are explored in parallel through column-cell communication. The reasoning process is enhanced by retrieval-augmented knowledge, and the final answer is refined through correction and summarization.}
\label{fig:MTQA}
\end{figure*}

This section presents the MoT framework in detail. As illustrated in Figure \ref{fig:MTQA}. The framework is designed to make fuller use of existing knowledge and to equip LLMs with both breadth and depth when exploring the problem’s solution space. The framework consists of two parts: (1) retrieval-augmentation, (2) Matrix of thought.

% Our method improves the reliability and comprehensiveness of LLM-generated answers from two aspects: adopting a superior knowledge representation to enhance the LLM’s comprehension. And employing a powerful thought structures to strengthen the LLM’s reasoning capability.

\subsection{Problem Definition}
A question-answering task comprises a given question \(Q\) and a set of input documents \(\mathcal{D} = \{ d_{1}, d_{2}, \ldots, d_{n} \}\), where \(n\) denotes the number of input documents. Both the question and the input documents are expressed in natural language. \(D\) is the auxiliary factual knowledge that supports the large language model \(L\) in answering \(Q\) and thus obtaining the answer \(A\). This factual knowledge can take the form of textual passages, triples in a knowledge graph, or Internet web page URLs. This paper focuses on treating text paragraphs as auxiliary factual knowledge. Using the retriever \(R\), we retrieve from \(D\) the factual knowledge set \(\mathcal{K} = \{ k_{1}, k_{2}, \ldots, k_{m} \}\) that matches the current input \(X\) to be verified by \(L\), where \(m\) represents the \text{top-}\(m\) text blocks ranked by their degree of match with \(X\).

In MoT, our objective is to construct a matrix of thought \(M\) that can carry out the widest and deepest possible inference within the solution space of question \(Q\), thereby identifying the optimal answer \(A\).

\subsection{Retrieval-Augmented}
\textbf{Step 1: Knowledge Database construction.}In MoT, the first step is to build a retrievable external knowledge base and partition the given set of documents into chunks whose size is suitable for processing by the LLM:

\begin{equation}
C = \{c_{1},\, c_{2},\, c_{3},\, \dots,\, c_{k}\},\quad
k = \left\lceil \frac{N_D}{C_{\text{len}}} \right\rceil
\label{eq:chunk-count}   
\end{equation}

Where \(C\) denotes the set of blocks after the document group \(D\) is chunked, \(k\) represents the number of blocks, \( N_D \) is the total number of words in the document group \(D\), and \( C_{len} \) is the length of the set text block.

Subsequently, we use LLM to recognize and extract entities (e.g., dates, places, people, events, etc.) and the relationships between them according to set cues (Detailed prompt are presented in the Appendix), generating a knowledge graph \(G\) that comprehensively encompasses the content of the document, which consists of the extracted entities \( v \) with their relationships with each other \( \varepsilon \), and reacts to the connections and focuses in the whole document set. 

First, llm embeds the chunked text block \( C_i \) after filling it to the prompt template:

\begin{equation}
L_{c_i} = \operatorname{embed}\!\bigl(\operatorname{template}(C_i, x)\bigr)
\end{equation}

Where \(x\) represents the template for performing entity and relationship extraction, \( C_i \) represents the text block, and \( L_{c_i}\) represents the generated embedding vectors subsequently input to the LLM for comprehension. The LLM performs recognition and extraction operations to extract the entity \( v \) and the relation \( \varepsilon \) from \( L_{c_i}\).

\begin{equation}
v,\; \varepsilon
  = \text{Extraction \& identification}\!\left(\sum_{i=1}^{k} L_{c_i}\right)
\end{equation}

Subsequently, we develop guidance llm to construct a key-value pair \((k, v)\) for each recognized and extracted entity \( v \) and the relation \( \varepsilon \) between them\cite{guo2024lightrag}. 
% The keys are composed of individual words or short phrases and are later used as query terms to retrieve the corresponding entities and relations. For entities, their names function as keys; for relations, the LLM summarizes the text that describes the link between two entities to generate multiple keys. The associated values are the source passages in the external knowledge base from which those entities and relations are drawn. The key–value mechanism both accelerates retrieval and preserves direct access to the source passages in the external knowledge base. During the generation of entities and their corresponding relations, the LLM augments each item with an explanatory description, allowing subsequent LLM-based reasoning modules to exploit this enriched knowledge more effectively for refinement and enhancement.

\begin{equation}
\hat{v},\; \hat{\varepsilon}
  = \operatorname{Deduplication}\!\bigl(v,\; \varepsilon\bigr)
\end{equation}

After completing the extraction of entity \( v \) and relation \( \varepsilon \), we perform a de-redundancy operation to identify and merge the duplicates of \( v \) and \( \varepsilon \) to obtain the unique \( \hat{v} \) and \( \hat{\varepsilon} \) in the whole knowledge graph.
\begin{equation}
G = \bigl(\hat{v},\; \hat{\varepsilon}\bigr)
\end{equation}
After completing the above operations, the entities \( \hat{v} \) as nodes and the relations \( \hat{\varepsilon} \) as edges construct the knowledge graph \(G\). Among them, the source entity \( \hat{v}_s \) and the target entity \( \hat{v}_t \) with the relation \( \hat{\varepsilon} \) between them form the basic knowledge graph structure: the knowledge graph triple\((\hat{v}_s, \hat{\varepsilon}, \hat{v}_t)\) , which serves as the basic unit for constructing the knowledge graph and retrieving the knowledge base.

After constructing the knowledge graph, we consider the incremental update operations for the knowledge graph when the document group is updated. For the newly added document group \( D' \), we follow the aforementioned steps to extract entities and relations, obtaining a new subgraph \(G' = (V', \varepsilon')\). Next, the entities and relations in \( G \) and \( G' \)are merged by taking their union, thereby combining \( G' \) with \( G \).

\textbf{Step 2: Retrieval knowledge through queries.} After constructing the external knowledge base, we build the retrieval function mechanism to match the knowledge that best supports the query. We use a dual retrieval mechanism \cite{edge2024local} to extract relevant information from the numerous and complex intertwined dependencies of entities in the external knowledge base. 

The retrieved knowledge consists of entities (nodes of the knowledge graph), relations (edges of the knowledge graph), and the original text (indexed by the key-value pairs of nodes and edges), forming knowledge units, as shown in Figure \ref{fig:MTQA}. The knowledge units simultaneously contain structured information, supporting multi-hop reasoning and emphasizing key information through knowledge graph triples\cite{saxena2020improving}, as well as undistorted, broader content original \cite{lewis2020retrieval} (such as complex metaphors, emotional expressions, etc.), which better aligns with the preferences of LLMs \cite{dong2025understand} in original text. Experimental results demonstrate that these knowledge units significantly enhance the reasoning performance of LLMs.

\subsection{Matrix of Thought}
\textbf{Step 3: Reasoning enhance.} Next, we start building the column units in MoT. A given query \( Q \) is fed into the retrieval enhancement module to get the knowledge unit \( KU_1 \) that underpins the reasoning of LLMs:

\begin{equation}
KU_1 = \text{Retrieval-Augmented}(S(Q))
\end{equation}

Next, we fill the knowledge unit \( KU_1 \) and the query \( Q \) into the prompt template, guiding the LLMs to leverage the information contained in the knowledge unit to enhance their understanding of the question. A strategy from shallow to deep is employed to guide the LLMs in generating the initial thought nodes:

\begin{equation}
Tn_1^1 = \text{LLM}(\text{template}(Q, KU_1))
\end{equation}

The initial nodes contain relatively shallow and indicative views and solutions to the question. MoT will iteratively explore and refine these nodes in terms of breadth and depth, gradually evolving them so that the generated thought nodes converge towards the target of the question.

\textbf{Step 4: Column cell communication mechanism.} To more efficiently explore the breadth of the problem and implement a multi-branching strategy, we have developed the column cell communication mechanism. Specifically, in MoT, a portion of the paragraph that represents the strategy and direction of the previous thought node in the same column is extracted (typically the last few paragraphs of the thought node). This portion is then used as an example to guide the LLMs in generating the next thought node, encouraging a completely different line of thought and strategy.

\begin{equation}
Tn_{m}^{n} = LLM\left(\mathrm{template}\left(Q, KU_{n}, \alpha Tn_{m-1}^{n}\right)\right)
\end{equation}

Where \( \alpha \) is the communication weight with range \( 0 \le \alpha \le 1 \). It denotes the proportion (rounded up) of paragraphs taken from node \( Tn_{m-1}^{\,n} \). A value of \( 0 \) means there is no communication between the two thought nodes, and a value of \( 1 \) means the entire content of \( Tn_{m-1}^{\,n} \) is passed to \( Tn_{m}^{\,n} \). This parameter determines the transparency of prior ideas to the LLMs. In MoT, the column-cell communication mechanism is essential and distinguishes MoT from ToT and related reasoning structures. Throughout MoT, we use a matrix \( A \) to represent the communication weights. Given an \( m \times n \) thought matrix, the weight matrix has dimension \((m-1)\times(n-1)\). The optimal value of this parameter is detailed in the Experiments section.

\begin{equation}
A =
\begin{bmatrix}
\alpha_{m-1}^{\,1} & \cdots & \alpha_{m-1}^{\,n-1} \\
\vdots             & \ddots & \vdots               \\
\alpha_{1}^{\,1}   & \cdots & \alpha_{1}^{\,n-1}
\end{bmatrix}
\end{equation}

\textbf{Step 5: Refinement and Summarization.}After generating the entire column of thought nodes, MoT matches all thought nodes within the current column cell to the knowledge units\( KU_{n}' \)in the external knowledge base, and uses the LLM together with \( KU_{n}' \) to refine and aggregate these thought nodes, yielding the summary node \( Sn^{\,n} \).

\begin{equation}
KU_{n}' = \text{Retrieval-Augmented}\!\left(S\!\left(Q, \sum_{i=1}^{m} Tn_{i}^{\,n}\right)\right)
\end{equation}

\begin{equation}
S_{n}^{\,n} = \text{LLM}\!\bigl(\text{template}(Q, KU_{n}')\bigr)
\end{equation}

The summary node corrects any fact-inconsistent strategies that appear among the thought nodes in the column, while preserving as much strategy diversity as possible. When generating the next column cell, the process builds on the current column’s summary node to further develop promising strategies. After a specified number of iterations, the final summary node is taken as the output of MoT. Algorithm\ref{alg:algorithm} in the appendix demonstrates the complete reasoning process of the MoT framework.

\subsection{Discussion}
In this chapter, we further elaborate on the objectives introduced in the Introduction. While controlling the framework’s runtime overhead, our method integrates external knowledge and refines it into knowledge units to help LLMs efficiently enhance reasoning and correct strategies. Moreover, by adopting a matrix structure and a column-cell communication mechanism, it encourages LLMs to proactively explore multiple dimensions of the solution space. It is worth noting that the MoT paradigm abstracts the LLM reasoning process as a matrix. When the number of columns or rows of this matrix equals 1, MoT degenerates into a CoT structure with RAG-based correction. When the column-cell communication weight is 0, MoT degenerates into a ToT-like structure. This indicates that CoT and ToT are special cases of MoT, which confers greater flexibility and adaptability on MoT.

\section{Experiment}
\subsection{Experimental Setup}
In this section, we design three experiments to validate the effectiveness of the MoT framework. (1) 24-Point Game: To verify the framework's logical and numerical reasoning capabilities. (2) Question Answering Evaluation: To assess the framework's ability to reduce hallucinations in complex problems. (3) Proposition Writing: To evaluate the framework's ability in multiple dimensions, such as the breadth and accuracy of responses when facing open-ended questions.

To implement the MoT framework, we default to GPT-4o-mini as the LLM \(L\). To ensure consistency with prior work \cite{trivedi2022interleaving}, the dataset chunk size is set to 1200. We use the nano vector database for vector database operations. The hyperparameter settings for MoT are detailed in the Numerical Analysis subsection of this chapter. All experiments are conducted on a Linux server equipped with six NVIDIA A6000 GPUs.

\subsection{24-Point Game}
The 24-Point Game is a classic mathematical puzzle task. The game rules are as follows: the player manipulates four integers, ranging from 1 to 9, using addition, subtraction, multiplication, and division operations to achieve a final result of 24. The order of operations can be flexibly adjusted as needed, and each of the four numbers can only be used once. This task serves as a benchmark for LLMs' mathematical reasoning abilities\cite{kim2024llm,zhangplanning}. It tests the LLM's arithmetic operations and strategic planning abilities, requiring exploration of various combinations and operation sequences to find the optimal solution. We tested the model on 100 randomly selected problems from the 4nums dataset. For each model, a result is labeled as successful if it includes an equation that complies with the rules. The baseline RATT, which uses the external knowledge base from RAG, is fully consistent with MoT.

\begin{table}[H]
\centering
\caption{Success rates of different methods.}
\label{tab:success_rates}
\begin{tabular}{lc}
\toprule
\textbf{Method} & \textbf{Success Rate} \\
\midrule
IO       & 0.17 \\
CoT      & 0.28 \\
ToT      & 0.36 \\
RAT      & 0.34 \\
RATT-RAG & 0.42 \\
RATT     & 0.45 \\
\midrule
MoT-RAG     & 0.52 \\
\textbf{MoT} & \textbf{0.66} \\
\bottomrule
\end{tabular}
\end{table}

As shown in Table \ref{tab:success_rates}, the success rate of MoT reached a maximum of 66\%, while the baseline methods, such as RATT, did not exceed 45\%. Interestingly, compared to the RATT method with the RAG module, our MoT-RAG method, which excludes the RAG module, still maintains a leading success rate, while the RATT-RAG method, without the RAG module, did not experience a significant drop in success rate. We believe that in logical and numerical reasoning tasks, the framework's own reasoning and thinking ability is key to solving the problem. If the reasoning ability is insufficient, even with sufficient input data, there remains the possibility of errors.

\subsection{Question Answering Evaluation}
In this experiment, we use NaturalQuestions \cite{kwiatkowski2019natural} for single-hop QA evaluation, and HotpotQA \cite{yang2018hotpotqa} together with 2WikiMultihopQA (2WMQA) \cite{ho2020constructing} for multi-hop QA evaluation. To ensure consistency with prior work, we sample 500 instances from each dataset. We use the F1 score and Exact Match (EM) as evaluation metrics for all three datasets. We compare MoT with GPT-4o-mini under direct prompting, Zero-shot CoT\cite{kojima2022large}, Few-shot CoT\cite{wei2022chain}, and ToT, and against RAG-based baselines including vanila RAG\cite{lewis2020retrieval}, ReAct\cite{yao2023react}, IRCoT\cite{trivedi2022interleaving}, FLARE\cite{jiang2023active}, Self-Rag\cite{asai2024self}, SearChain\cite{xu2024search}, Rowen\cite{ding2024retrieve}, SlimPLM\cite{tan2024small}, LightRAG\cite{guo2024lightrag}, CoA\cite{zhang2024chain}, RAT\cite{wang2024rat}, and RATT\cite{zhang2025ratt}. For RATT, we set its internal RAG knowledge base to be the same as that of MoT.

\begin{table*}[ht!]
\centering
% \resizebox{\textwidth}{!}{
\begin{tabular}{@{}c|cc|cc|cc@{}}
\toprule
\textbf{Dataset}
& \multicolumn{2}{c|}{NQ}
& \multicolumn{2}{c|}{HotpotQA}
& \multicolumn{2}{c}{2WMQA} \\
\cmidrule(lr){2-3} \cmidrule(lr){4-5} \cmidrule(l){6-7}
\textbf{Metric}& F1 & EM & F1 & EM & F1 & EM \\ \midrule
Vanilla GPT 4o-mini  & 0.425 & 0.298 & 0.380 & 0.265 & 0.316 & 0.229 \\
Zero-shot CoT        & 0.457 & 0.295 & 0.355 & 0.262 & 0.324 & 0.216 \\
Few-shot CoT         & 0.447 & 0.298 & 0.376 & 0.255 & 0.363 & 0.227 \\
ToT                  & 0.456 & 0.305 & 0.368 & 0.267 & 0.345 & 0.237 \\ \midrule
Vanilla RAG          & 0.385 & 0.258 & 0.387 & 0.254 & 0.314 & 0.244 \\
ReAct                & 0.335 & 0.212 & 0.390 & 0.270 & 0.305 & 0.204 \\
IRCoT                & 0.344 & 0.216 & 0.361 & 0.232 & 0.318 & 0.202 \\
FLARE                & 0.455 & 0.318 & 0.391 & 0.268 & 0.364 & 0.246 \\
Self-Rag             & 0.387 & 0.270 & 0.357 & 0.220 & 0.311 & 0.210 \\
SearChain            & 0.337 & 0.214 & 0.349 & 0.216 & 0.313 & 0.222 \\
Rowen                & 0.452 & 0.286 & 0.382 & 0.240 & 0.307 & 0.212 \\
SlimPLM              & 0.442 & 0.280 & 0.393 & 0.266 & 0.368 & 0.242 \\ 
LightRAG             & 0.439 & 0.312 & 0.362 & \underline{0.282} & 0.356 & 0.276 \\
CoA                  & 0.452 & 0.326 & 0.387 & 0.272 & 0.362 & 0.243 \\
RAT                  & 0.462 & \underline{0.338} & 0.392 & 0.275 & 0.376 & 0.239 \\
RATT                 & \underline{0.479} & 0.323 & \underline{0.410} & 0.279 & \underline{0.406} & \underline{0.284} \\ \midrule
MoT & \hspace{5.7mm}\textbf{0.510}\textsuperscript{\scalebox{0.8}{\textcolor{red}{\(\triangle\)}3.1\%}}
     & \hspace{5.7mm}\textbf{0.368}\textsuperscript{\scalebox{0.8}{\textcolor{red}{\(\triangle\)}3.0\%}}
     & \hspace{5.7mm}\textbf{0.452}\textsuperscript{\scalebox{0.8}{\textcolor{red}{\(\triangle\)}4.2\%}}
     & \hspace{5.7mm}\textbf{0.318}\textsuperscript{\scalebox{0.8}{\textcolor{red}{\(\triangle\)}3.6\%}}
     & \hspace{5.7mm}\textbf{0.454}\textsuperscript{\scalebox{0.8}{\textcolor{red}{\(\triangle\)}4.8\%}}
     & \hspace{5.7mm}\textbf{0.326}\textsuperscript{\scalebox{0.8}{\textcolor{red}{\(\triangle\)}4.2\%}} \\ \midrule
\end{tabular}
% }
\caption{Overall results on three datasets for multi-metric performance evaluation. The existing best results are underlined and the best results are in bold.\textsuperscript{\scalebox{0.8}{\textcolor{red}{\(\triangle\)}}} indicates the improvement of our results over the state-of-the-art results.}
\vspace{-0.3cm}
\label{tab:f1_em_results}
\end{table*}

\textbf{Results Analysis:} We compared MoT with the aforementioned baselines, as shown in Table 1. Based on the comparison metrics, we reached the following conclusions: First, MoT outperforms all baselines. Compared to non-RAG baseline methods, our approach improves F1 by 5.3\% to 9.1\% and EM by 6.2\% to 8.9\%. Furthermore, when compared to RAG-based and thought-structure-based methods, we still maintain a lead of at least 3\% in both F1 and EM scores. This is particularly evident on multi-hop question-answering datasets such as HotpotQA and 2WMQA, demonstrating the effectiveness of the proposed MoT paradigm and knowledge units. The column-cell communication mechanism of MoT focuses on exploring multi-branch thinking and enhancing the breadth of responses to optimize answer quality. Second, we found that some RAG-based methods did not fully outperform the non-RAG baselines, possibly due to inefficient retrieval mechanisms that introduce noise or erroneous information, misleading the LLM’s responses.

\subsection{Proposition Writing}
To address the issues in Multi-metric performance evaluation, we build on the method of Guo et al. \cite{edge2024local} and design a simple yet effective protocol to evaluate, in a multidimensional manner, the response quality of QA models for open-domain QA tasks. We selected datasets from four domains—biology, law, computer science, and physics—from the UltraDomain dataset\cite{qian2025memorag} to construct the external knowledge base, with each dataset containing millions of tokens. Following the method of Edge et al\cite{edge2024local}, we retrieved open-ended questions from the relevant domains. For each question, we guided GPT-4o to evaluate MoT and the baseline models in four dimensions: Comprehensiveness, Accuracy, Empowerment, and Overall performance. The results were used to determine the winning rate and provide the rationale for the judgment.

\begin{table*}[ht]
\centering
\resizebox{\textwidth}{!}{
\begin{tabular}{lcc|cc|cc|cc}
\toprule
\multirow{2}{*}{\textbf{Metric}} & \multicolumn{2}{c|}{\textbf{Biology}} & \multicolumn{2}{c|}{\textbf{CS}} & \multicolumn{2}{c|}{\textbf{Legal}} & \multicolumn{2}{c}{\textbf{Physics}} \\ \cline{2-9}  \addlinespace[0.5ex]
                                 & \textbf{LightRAG} & \textbf{MoT} & \textbf{LightRAG} & \textbf{MoT} & \textbf{LightRAG} & \textbf{MoT} & \textbf{LightRAG} & \textbf{MoT} \\ \hline
\textbf{Comprehensiveness}       & 5.6\%  & \textbf{94.4\%}  & 4.8\% & \textbf{95.2\%} & 8.8\% & \textbf{91.2\%} & 5.2\% & \textbf{94.8\%} \\ 
\textbf{Accuracy}                & 6.4\%  & \textbf{93.6\%} & 4.0\% & \textbf{96.0\%} & 4.8\% & \textbf{95.2\%} & 5.6\% & \textbf{94.4\%} \\ 
\textbf{Empowerment}             & 3.6\%  & \textbf{96.4\%} & 6.8\% & \textbf{93.2\%} & 6.0\%  & \textbf{94.0\%} & 7.6\% & \textbf{92.4\%} \\ 
\textbf{Overall}                 & 4.4\%  & \textbf{95.6\%} & 3.6\% & \textbf{96.4\%} & 5.2\% & \textbf{94.8\%} & 5.2\% & \textbf{94.8\%} \\ \cline{2-9}  \addlinespace[0.5ex]
                                 & \textbf{CoA} & \textbf{MoT} & \textbf{CoA} & \textbf{MoT} & \textbf{CoA} & \textbf{MoT} & \textbf{CoA} & \textbf{MoT} \\ \midrule
\textbf{Comprehensiveness}       & 13.6\% & \textbf{86.4\%} & 9.6\%  & \textbf{90.6\%} & 11.2\% & \textbf{88.8\%} & 17.2\% & \textbf{80.8\%} \\ 
\textbf{Accuracy}                & 15.6\% & \textbf{84.4\%} & 13.6\% & \textbf{86.4\%} & 13.6\% & \textbf{86.4\%} & 12.0\% & \textbf{88.0\%} \\ 
\textbf{Empowerment}             & 17.2\% & \textbf{82.8\%} & 15.6\% & \textbf{84.8\%} & 16.4\% & \textbf{83.6\%} & 18.8\% & \textbf{81.2\%} \\ 
\textbf{Overall}                 & 15.2\% & \textbf{84.8\%} & 10.4\% & \textbf{89.6\%} & 12.4\% & \textbf{87.6\%} & 13.2\% & \textbf{86.8\%} \\ \cline{2-9}  \addlinespace[0.5ex]
                                 & \textbf{RAT} & \textbf{MoT} & \textbf{RAT} & \textbf{MoT} & \textbf{RAT} & \textbf{MoT} & \textbf{RAT} & \textbf{MoT} \\ \midrule
\textbf{Comprehensiveness}       & 17.2\% & \textbf{82.8\%} & 10.8\% & \textbf{89.2\%} & 16.4\% & \textbf{83.6}\% & 18.8\% & \textbf{81.2\%} \\ 
\textbf{Accuracy}               & 17.6\%  & \textbf{82.4\%} & 25.2\% & \textbf{74.8\%} & 23.6\% & \textbf{76.4\%} & 22.4\% & \textbf{77.6\%} \\ 
\textbf{Empowerment}             & 19.6\% & \textbf{80.4\%} & 22.8\% & \textbf{77.2\%} & 16.4\% & \textbf{83.6\%} & 20.8\% & \textbf{79.2\%} \\ 
\textbf{Overall}                 & 16.4\% & \textbf{83.6\%} & 11.2\% & \textbf{88.8\%} & 15.2\% & \textbf{84.8\%} & 18.4\% & \textbf{81.6\%} \\ \cline{2-9}  \addlinespace[0.5ex]
                                 & \textbf{RATT} & \textbf{MoT} & \textbf{RATT} & \textbf{MoT} & \textbf{RATT} & \textbf{MoT} & \textbf{RATT} & \textbf{MoT} \\ \midrule
\textbf{Comprehensiveness}       & 16.8\% & \textbf{83.2\%} & 12.0\% & \textbf{88.0\%} & 22.4\% & \textbf{77.6\%} & 29.6\% & \textbf{70.4\%} \\ 
\textbf{Accuracy}                & 34.4\% & \textbf{65.6\%} & 32.0\% & \textbf{68.0\%} & 16.4\% & \textbf{83.6\%} & 32.0\% & \textbf{68.0\%} \\ 
\textbf{Empowerment}             & 24.0\% & \textbf{76.0\%} & 19.2\% & \textbf{80.8\%} & 24.8\% & \textbf{75.2\%} & 19.2\% & \textbf{80.8\%} \\ 
\textbf{Overall}                 & 18.4\% & \textbf{81.6\%} & 15.2\% & \textbf{84.8\%} & 18.4\% & \textbf{81.6\%} & 15.2\% & \textbf{84.8\%} \\ \midrule

\end{tabular}
}
\caption{The winning rate of MoT and its baseline in four datasets and four evaluation dimensions in the Proposition Writing. The one with the higher winning rate is marked in bold font size.}
\vspace{-0.5cm}
\label{tab:win_results}
\end{table*}

\textbf{Results Analysis:} As shown in Table \ref{tab:win_results}, we compare the win rates between MoT and the aforementioned baselines pairwise. It is evident that our method achieves a win rate of over 80\% across most dimensions and in the overall assessment, further validating the effectiveness of the MoT approach. Among them, the non-reasoning-structure method (LightRAG) achieves a win rate of no more than 10\% across the four datasets. This further supports our viewpoint that, under similar knowledge enhancement, reasoning structures play a critical role in the utilization of knowledge and the logical reasoning ability, significantly influencing the overall quality of the answers. The chain-based reasoning structure methods (CoA, RAT) achieve win rates ranging from 9.6\% to 25.2\%, with the win rate for comprehensiveness assessment consistently lower than that of other dimensions. This indicates that while chain-based reasoning structures have a positive effect on answer generation, their lack of multi-branch reasoning capability results in a noticeable disadvantage in comprehensiveness compared to tree-based structures (e.g., RATT) and matrix-based structures (e.g., MoT). The win rates for the tree-based reasoning structure method (RATT) range from 12\% to 34.4\%. We attribute this advantage to the column-cell communication mechanism employed by MoT, which better stimulates the LLMs' multi-branch reasoning strategies for answering the questions, thereby yielding higher-quality answers.

\subsection{Ablation Study}
Subsequently, we conducted an ablation study and developed four variants of MoT to demonstrate the effectiveness of each mechanism within MoT. The comparison is performed using the single-hop QA dataset NQ and the multi-hop QA dataset HotpotQA. The experimental results are shown in Table \ref{tab:ablation_results}.

The "-RAG" variant removes the RAG correction module, resulting in significant performance degradation on two datasets due to knowledge boundaries and LLM hallucinations. However, it outperforms non-RAG methods, demonstrating the superiority of the Matrix of Thought (MoT) over other thought structures. The "-origin" variant removes the use of original text in knowledge units, leading to some performance loss, especially on multi-hop datasets, where the performance gap between "-origin" and "-KG" is more pronounced, highlighting the advantage of KG-based data in multi-hop tasks. The "-KG" variant removes knowledge graph triples from the knowledge unit, showing noticeable performance loss due to the lack of structured multi-hop information in the retrieval process. The "-comm" variant, which disables column-cell communication by setting the communication weight matrix to 0, effectively reduces MoT to a tree structure. The results show that removing the communication mechanism leads to substantial performance loss, even exceeding the performance decline seen in the "-KG" and "-origin" variants, emphasizing the dominant role of the thought structure in LLM reasoning, beyond just enhanced knowledge.

\begin{table}[ht]
\centering
\setlength{\tabcolsep}{3mm}
\begin{tabular}{ccc|cc}
\toprule
\multirow{2}{*}{\vspace{-1.5mm}{\textbf{\makecell{Dataset \\ Metric}}}} & \multicolumn{2}{c|}{NQ} & \multicolumn{2}{c}{HotpotQA} \\ \cmidrule{2-3} \cmidrule{4-5} 
 & F1 & EM  & F1 & EM \\ \midrule
-RAG       & 0.467        & 0.318  & 0.384        & 0.278        \\
-origin    & 0.489        & 0.334  & 0.423        & 0.294        \\
-KG        & 0.483        & 0.329  & 0.407        & 0.273       \\
-comm      & 0.474        & 0.327  & 0.409        & 0.275        \\ \midrule
MoT       & 0.510        & 0.368  & 0.452        & 0.318        \\ \midrule
\end{tabular}
\caption{Comparison of F1 and EM scores of MoT and its four variant models on two datasets.}
\label{tab:ablation_results}
\vspace{-1cm}
\end{table}

\subsection{Numerical Analysis}
% In this subsection, we investigate two crucial hyperparameters in the MoT framework: the column-cell communication weight matrix and the size of the thought matrix. We explore their impact on both the performance and efficiency of the QA task using the HotpotQA dataset. The appendix also analyzes the joint impact of these two parameters.

\subsubsection{Weight Matrix Analysis}
First, we discuss the impact of the column-cell communication weight matrix settings on the framework's performance. In this experiment, we conduct a comparative study of 12 different weight matrix configuration methods using a \(3\times4\) matrix as an example. The configuration methods include Uniform, Gaussian, four different constant matrices (0.2, 0.5, 0.8, 1.0), as well as vertically increasing matrices (Vert-0.1, Vert-0.2, incrementing row by row along the y-axis), horizontally increasing matrices (Hor-0.1, Hor-0.2, incrementing column by column along the x-axis), and combined vertical and horizontal increasing matrices (Vert\&Hor-0.1, Vert\&Hor-0.2, incrementing both row by row and column by column along the x- and y-axes). The Gaussian distribution is set with a mean of 0 and a variance of 1, and the generated matrix is normalized to ensure each weight is between 0 and 1. The evaluation metrics for the experiment are the F1 score and Exact Match (EM) score.

\begin{figure}[htbp]
\centering
%\vspace{-1.2cm}
\includegraphics[width=1\columnwidth]{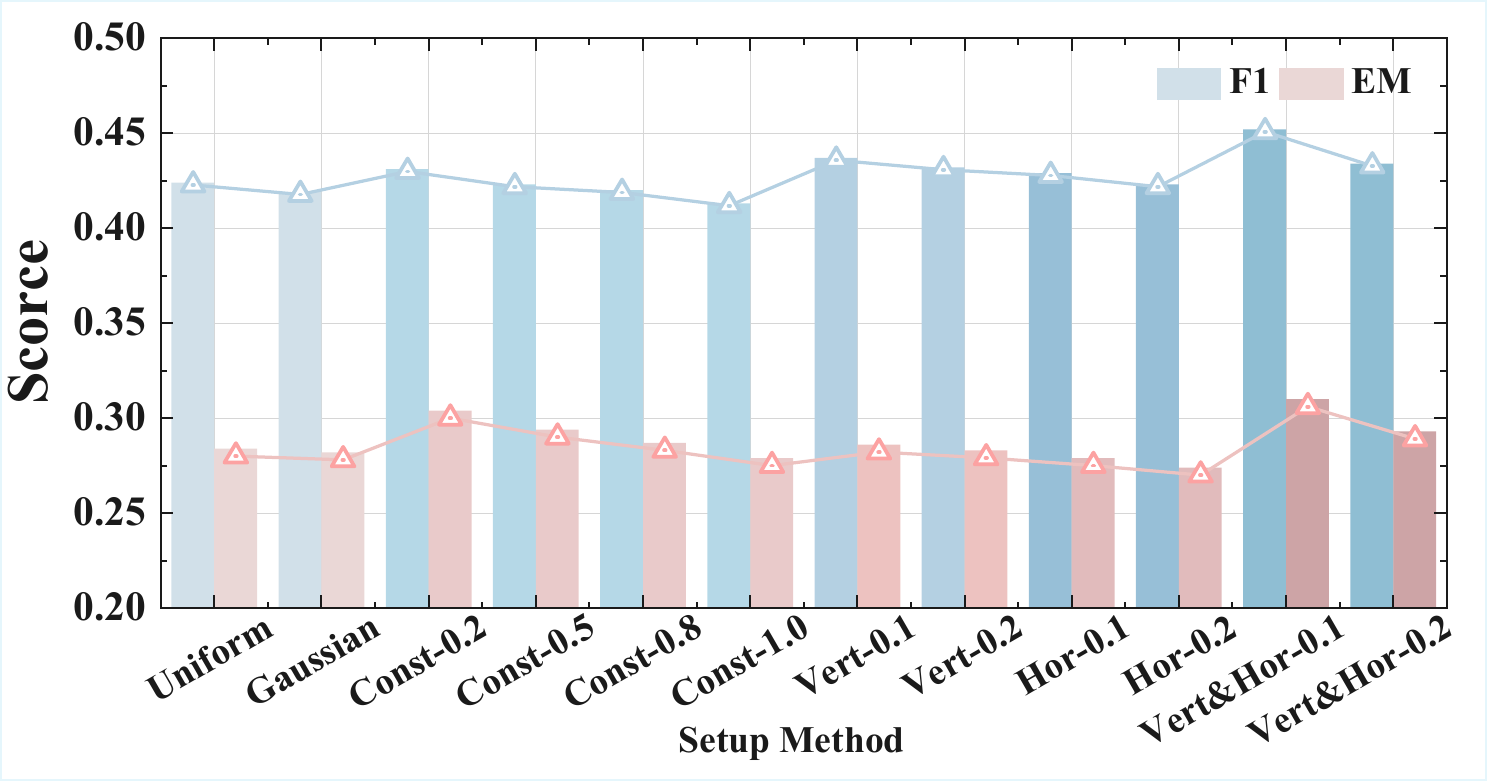}
% \vspace{-0.5cm}
\caption{Comparison of F1 and EM scores for various weight setting methods.}
\vspace{-0.5cm}
\label{fig:weight}
\end{figure}

The experimental results are shown in Figure \ref{fig:weight}. We observe that the Vert\&Hor-0.1 weight setting method achieves the best performance among all methods, with F1 and EM scores of 0.452 and 0.310, respectively. However, the performance of Vert\&Hor-0.2 decreases instead. We infer the following reason: In the early stages of reasoning, the model tends to passively adopt the thought process from the previous thought node, resulting in strategies that are highly similar to the previous ones. Therefore, it is necessary to suppress the influence of the previous thought node's reasoning in the early stages to further stimulate the LLMs' reasoning abilities and explore different strategies. The performance of the constant matrix with a weight of 0.2 outperforms that of matrices with weights of 0.5, 0.8, and 1.0, further supporting this viewpoint. In the middle and later stages of reasoning, MoT has already conducted a deep exploration of the solution space for the problem. However, it may forget the reasoning process and conclusions from the earlier thought nodes, potentially generating strategies that are repetitive. Therefore, it is necessary to expose more conclusions and thought processes, allowing the LLM to actively avoid generating strategies that have already been produced.

\subsubsection{MoT Size Analysis}
We next discuss the impact of MoT size on its performance and computational overhead.
Specifically, we examine how different matrix sizes affect MoT’s reasoning performance
(F1 score and EM) and its time cost (Time). The considered matrix sizes include six
configurations: \(2\times2\), \(2\times3\), \(3\times3\), \(3\times4\), \(4\times4\), and \(4\times5\).
The weight matrix for each configuration is set using the Vert\&Hor-0.1 method described above.

\begin{figure}[htbp]
\centering
\vspace{-0.7cm}
\includegraphics[width=\columnwidth]{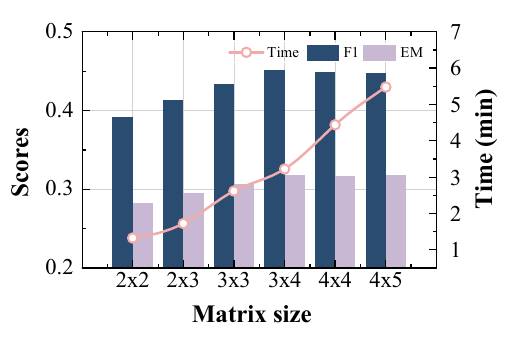}
\vspace{-1cm}
\caption{Comparison of F1 and EM scores for different matrix sizes.}
\vspace{-0.5cm}
\label{fig:size}
\end{figure}

As shown in Figure \ref{fig:size},experimental results indicate that as the matrix size increases, both the F1 and EM scores gradually improve. Specifically, the F1 score increases from 0.392 for a \(2\times2\) matrix to 0.452 for a \(3\times4\) matrix, while larger matrices stabilize around 0.45, with a slight decline. This suggests that smaller matrices may not explore the solution space sufficiently for typical multi-hop problems, making larger matrices effective for enhancing reasoning performance. However, this approach is not absolute—beyond a certain size, the strategies supported by existing knowledge are fully explored, and further exploration may lead to hallucinations and inaccurate answers. Additionally, reasoning time increases significantly with matrix size, from 1.324 minutes for a \(2\times2\) matrix to 5.483 minutes for a \(4\times5\) matrix. Considering both performance and computational time, the \(3\times4\) matrix offers the best balance, with optimal F1 and EM scores and relatively short reasoning time. For highly complex problems involving extensive knowledge, larger matrices may be considered.

Additionally, the combined effect of the two parameters is analyzed in the appendix.

% \begin{figure}[htbp]
% \centering
% \vspace{-0.5cm}
% \includegraphics[width=\columnwidth]{AnonymousSubmission/LaTeX/hot.pdf}
% \caption{Comparison of MTQA performance under different matrix sizes and different weight setting methods.}
% \label{fig:hot}
% \end{figure}

% \subsubsection{Joint Analysis}
% After obtaining preliminary conclusions for the weight matrix and MoT size individually,
% we further investigate the interaction between these two hyperparameters within the same
% experimental framework. The experimental results, shown in Figure \ref{fig:hot}, reveal a pattern in the heatmap that aligns with the previously observed trends. Based on this, we reached the following conclusions:

% (i) the careful design of the weight matrix is more beneficial than blindly
% enlarging the thought matrix, with small-step vertical--horizontal increments effectively
% activating diverse reasoning within a column without excessively leaking historical information;

% (ii) the thought matrix size has a ``sweet spot''—a sufficient number of rows and columns
% ensures adequate breadth and depth in strategy exploration, but beyond \(3\times 4\) the
% marginal gains diminish while time costs rise sharply. Based on these findings, we adopt
% \(3\times 4\) + Vert\&Hor-0.1 as the default MTQA configuration in all comparative experiments above.

\subsubsection{Time Complexity Analysis}
We compared the time overhead of MoT with the best baseline, RATT, on the UltraDomain-Biology and HotpotQA datasets, and provided the theoretical proof of the time overhead for both under the same size in the appendix.

\begin{table}[htbp]
\centering
\setlength{\tabcolsep}{1.5mm}
\begin{tabular}{ccc|cc}
\toprule
\multirow{2}{*}{\vspace{-1.5mm}{\textbf{\makecell{Method \\ Dataset}}}} & \multicolumn{2}{c|}{\textbf{MoT}} & \multicolumn{2}{c}{\textbf{RATT}} \\
\cmidrule{2-3} \cmidrule{4-5} 
& Biology & HotpotOA & Biology & HotpotOA \\
\midrule
125 times & 402.94 & 394.56 & 2804.25 & 2743.58 \\
average & 3.22 & 3.16 & 22.43 & 21.95 \\
\bottomrule
\end{tabular}
\caption{Comparison of time costs between MoT and RATT.}
\vspace{-0.5cm}
\label{tab:MTQA&RATT}
\end{table}

As shown in Table \ref{tab:MTQA&RATT}, both methods were tested 125 times on the Biology dataset.
The average reasoning time for MoT was 3.22 minutes, while RATT required 22.43 minutes.
Similarly, on the HotpotQA dataset, the average reasoning time for MoT was 3.16 minutes,
while RATT required 21.95 minutes. The reasoning time of MoT is approximately one-seventh
of that of RATT. This demonstrates that the MoT method significantly reduces computational
overhead and provides efficient reasoning capabilities when handling complex question-answering tasks.

% Next, we continue to analyze the time complexity composition of the two methods.
% Let the constructed MTQA have \( m \) rows and \( n \) columns, and correspondingly,
% the RATT tree has \( n \) layers, each with \( m \) nodes. Let \( k \) be the average time
% to call the LLM to get a response, and \( r \) be the time to call RAG. Then:

% \begin{equation}
% T_{RATT} = O(n(2m+1)k) + O(mnr) = O(n((2m+1)k + mr))
% \end{equation}

% \begin{equation}
% T_{MTQA} = O(mnk) + O(2nr) = O(n(mk + 2r))
% \end{equation}

% In RATT, each layer involves calling the LLM \( m \) times to generate thought nodes, 
% followed by \( m \) calls to the LLM and \( m \) calls to RAG to refine the thought nodes. 
% Finally, a single LLM call is made to generate the summary node. In contrast, in MTQA, 
% we simplify this process by calling RAG once to construct the knowledge unit, followed by 
% \( m \) LLM calls to generate the thought nodes enhanced by the knowledge unit. The process 
% is then completed by one call to RAG and one call to LLM for refinement and summarization. 
% Both experimental and theoretical results demonstrate the high efficiency and performance 
% of our approach.

\section{Conclusion}
In this paper, we propose a novel MoT reasoning paradigm that embeds knowledge units into the retrieval-augmented process. Through the column-cell communication mechanism, it triggers multi-branch thinking horizontally and accumulates deep strategies vertically, overcoming the common limitations of CoT's singular path and ToT's redundancy within the same layer. We conducted extensive experiments demonstrating the advantages over existing methods. Additionally, we reduced the reasoning time to one-seventh of the best baseline, balancing both accuracy and efficiency. This advantage is particularly evident in open and abstract tasks, such as high-level contextual understanding.

% \textbf{Limitations:} Although MTQA and the core MoT paradigm perform excellently on complex question-answering tasks, there are still several limitations that require further exploration. These include the fact that the effectiveness of the reasoning process and the final responses still depend on the quality of task prompts and the external knowledge base, as well as the high computational demands and the need for manually configured hyperparameters. For future work, we plan to extend the MoT paradigm to various LLM reasoning and alignment tasks, incorporating multimodal information and adaptive communication weight matrices. We aim to validate its generalizability and robustness in broader application scenarios and explore more suitable RAG systems for reasoning structures to achieve higher-quality reasoning enhancement and knowledge correction.

\bibliographystyle{ACM-Reference-Format}
\bibliography{sample-base}

\appendix

\section{Algorithm}
\begin{algorithm}[H]
\caption{MoT Framework}
\label{alg:MTQA}
\begin{algorithmic}[1]
\REQUIRE Question \( Q \), Document set \( \mathcal{D} = \{ d_1, d_2, \dots, d_n \} \), Retriever \( R \), LLM \( L \), Knowledge graph \( G \)
\ENSURE Final Answer \( A \)

\STATE \textbf{Step 1: Knowledge Database Construction}
\STATE \( C \gets \{ c_1, c_2, \dots, c_k \} \)
\FOR{each chunk \( c_i \in C \)}
    \STATE \( L_{c_i} \gets \operatorname{embed}(\operatorname{template}(c_i, x)) \)
    \STATE \( v, \varepsilon \gets \text{Extraction \& Identification}(L_{c_i}) \)
\ENDFOR
\STATE \( \hat{v}, \hat{\varepsilon} \gets \operatorname{Deduplication}(v, \varepsilon) \)
\STATE \( G \gets (\hat{v}, \hat{\varepsilon}) \)

\textbf{Step 2: Retrieval via Queries}
\STATE \( k^{(d)}, k^{(g)} \gets \text{Extract keywords}(Q) \)
\STATE \( \mathcal{K} \gets \operatorname{Retrieve}(k^{(d)}, k^{(g)}, G) \)
\STATE \( KU \gets \{ \mathcal{K}_\text{Triples} + \mathcal{K}_\text{Source Text} \} \)

\textbf{Step 3: Initial Thought Node Generation}
\STATE \( KU_1 \gets \text{Retrieval-Augmented}(S(Q)) \)
\STATE \( Tn_1^1 \gets L(\operatorname{template}(Q, KU_1)) \)

\textbf{Step 4: Column Cell Communication}
\FOR{each column \( n \), row \( m \)}
    \STATE \( Tn_m^n \gets L(\operatorname{template}(Q, KU_n, \alpha Tn_{m-1}^n)) \)
    \STATE \( A \in \mathbb{R}^{(m-1)\times(n-1)} \)
\ENDFOR

\textbf{Step 5: Refinement and Summarization}
\STATE \( KU_n' \gets \text{Retrieval-Augmented}(S(Q, \sum_{i=1}^m Tn_i^n)) \)
\STATE \( S_n^n \gets L(\operatorname{template}(Q, KU_n')) \)

\textbf{Step 6: Output}
\STATE \( A \gets S_{n_{\text{final}}}^{\,n_{\text{final}}} \)
\end{algorithmic}
\label{alg:algorithm}
\end{algorithm}

\section{Theoretical proof of time complexity}
Let the constructed MoT have \( m \) rows and \( n \) columns, and correspondingly,
the RATT tree has \( n \) layers, each with \( m \) nodes. Let \( k \) be the average time
to call the LLM to get a response, and \( r \) be the time to call RAG. Then:

\begin{equation}
T_{RATT} = O(n(2m+1)k) + O(mnr) = O(n((2m+1)k + mr))
\end{equation}

\begin{equation}
T_{MoT} = O(mnk) + O(2nr) = O(n(mk + 2r))
\end{equation}

In RATT, each layer involves calling the LLM \( m \) times to generate thought nodes, 
followed by \( m \) calls to the LLM and \( m \) calls to RAG to refine the thought nodes. 
Finally, a single LLM call is made to generate the summary node. In contrast, in MoT, 
we simplify this process by calling RAG once to construct the knowledge unit, followed by 
\( m \) LLM calls to generate the thought nodes enhanced by the knowledge unit. The process 
is then completed by one call to RAG and one call to LLM for refinement and summarization. 
Both experimental and theoretical results demonstrate the high efficiency and performance 
of our approach.

\section{Joint Numerical Analysis}
\begin{figure*}[htbp]
\centering
\vspace{-0.5cm}
\includegraphics[width=2\columnwidth]{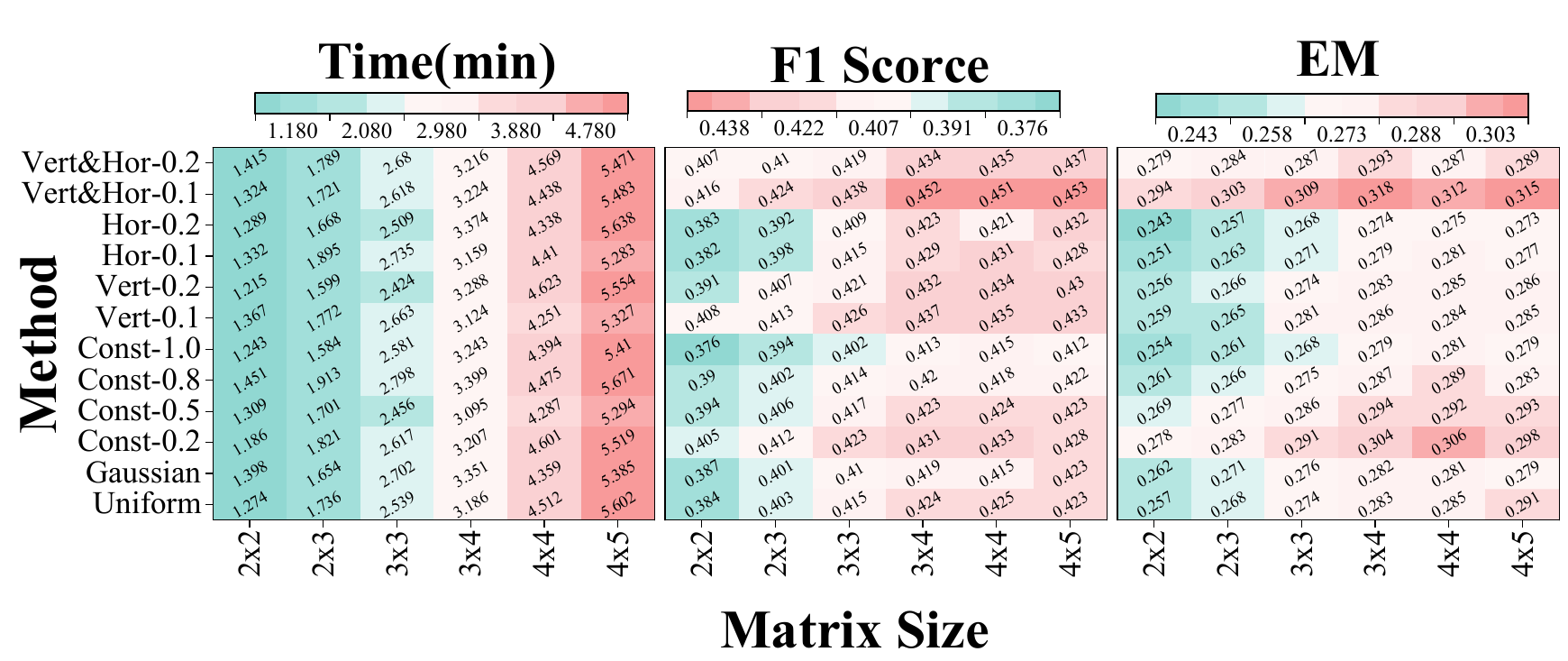}
\caption{Comparison of MoT performance under different matrix sizes and different weight setting methods.}
\label{fig:hot}
\end{figure*}

After obtaining preliminary conclusions for the weight matrix and MoT size individually,
we further investigate the interaction between these two hyperparameters within the same
experimental framework. The experimental results, shown in Figure \ref{fig:hot}, reveal a pattern in the heatmap that aligns with the previously observed trends. Based on this, we reached the following conclusions:

(i) the careful design of the weight matrix is more beneficial than blindly
enlarging the thought matrix, with small-step vertical--horizontal increments effectively
activating diverse reasoning within a column without excessively leaking historical information;

(ii) the thought matrix size has a ``sweet spot''—a sufficient number of rows and columns
ensures adequate breadth and depth in strategy exploration, but beyond \(3\times 4\) the
marginal gains diminish while time costs rise sharply. Based on these findings, we adopt
\(3\times 4\) + Vert\&Hor-0.1 as the default MoT configuration in all comparative experiments above.

\section{Example of Proposition Writing}
\begin{figure}[htbp]
\centering
%\vspace{-1.2cm}
\includegraphics[width=\columnwidth]{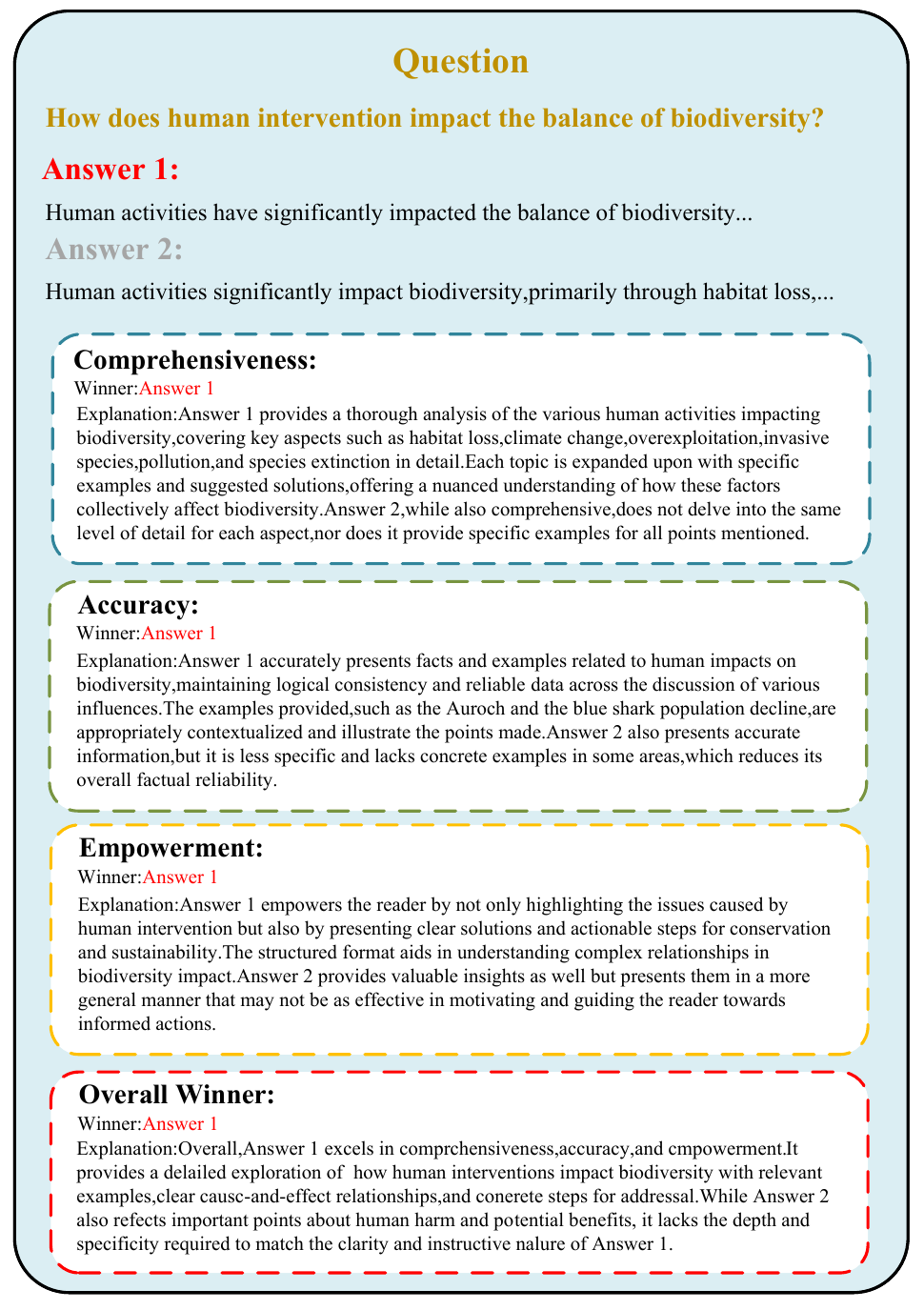}
\caption{example.}
\label{fig:example}
\end{figure}

GPT-4o provides an evaluation of proposition writing, identifying the winning side and the reasons for its victory, allowing for manual verification of the evaluation's validity. The entire evaluation process is shown in Figure \ref{fig:example}.

\section{Main prompt template}
In this section, we present various LLM data processing prompt templates mentioned in the Methodology section, including Figures \ref{fig:Entity_Relationship_Extraction}, \ref{fig:Extract_key}, \ref{fig:Initial_Thought_Node}, and \ref{fig:Summary_Node}.

\subsection{Prompt for graph construct}
\begin{figure}[htbp]
\centering
%\vspace{-1.2cm}
\includegraphics[width=\columnwidth]{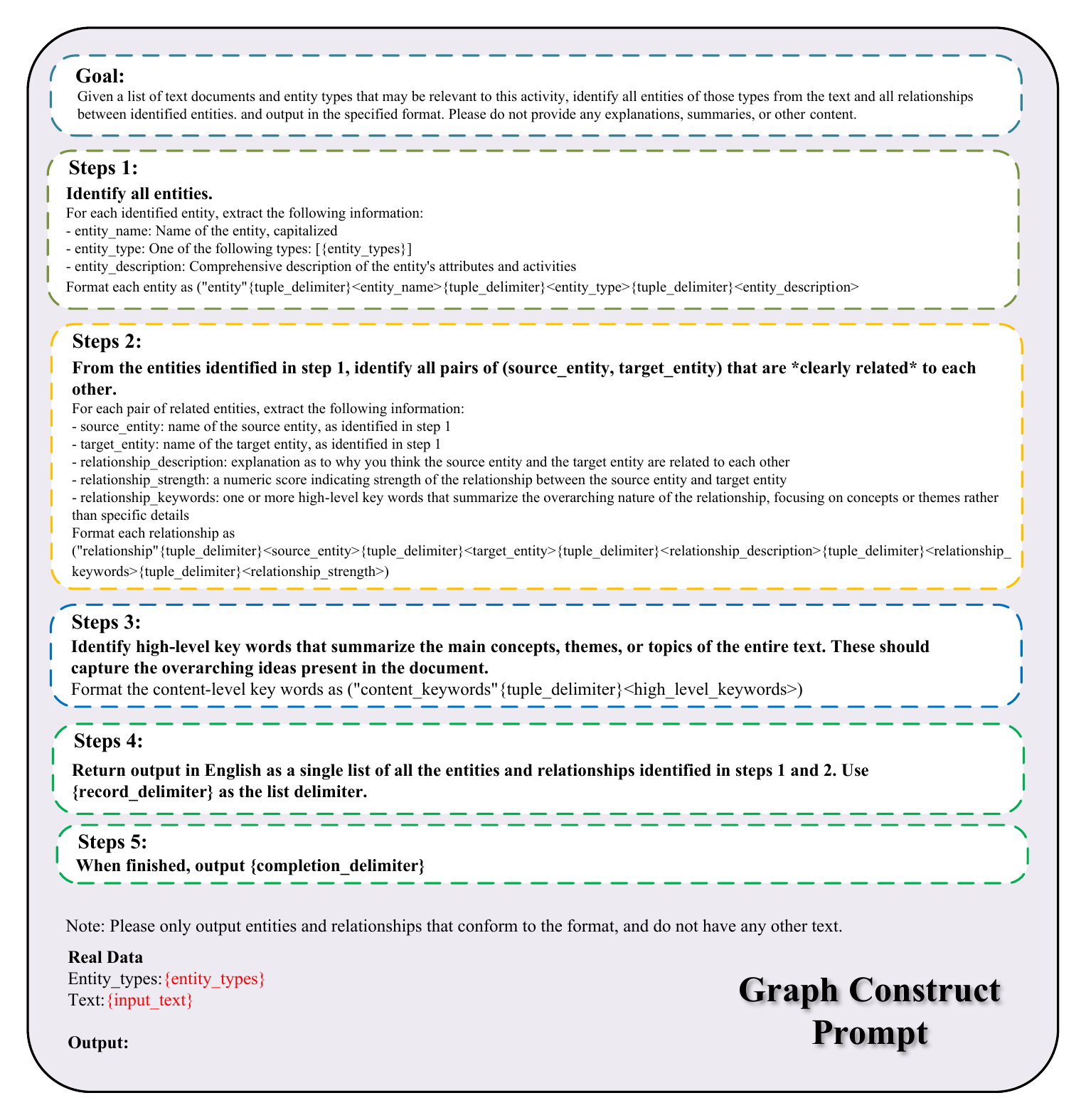}
\caption{Graph Construct Prompt.}
\label{fig:Entity_Relationship_Extraction}
\end{figure}

As shown in Figure \ref{fig:Entity_Relationship_Extraction}, the graph construction prompt template is used to build a knowledge graph from text. It consists of several steps: first identifying entities and their descriptions, then extracting relationships between entities, followed by capturing high-level thematic keywords, and finally outputting everything in a structured list. It emphasizes strict formatting without additional explanations or summaries.

\subsection{Prompt for extract key}
\begin{figure}[htbp]
\centering
%\vspace{-1.2cm}
\includegraphics[width=\columnwidth]{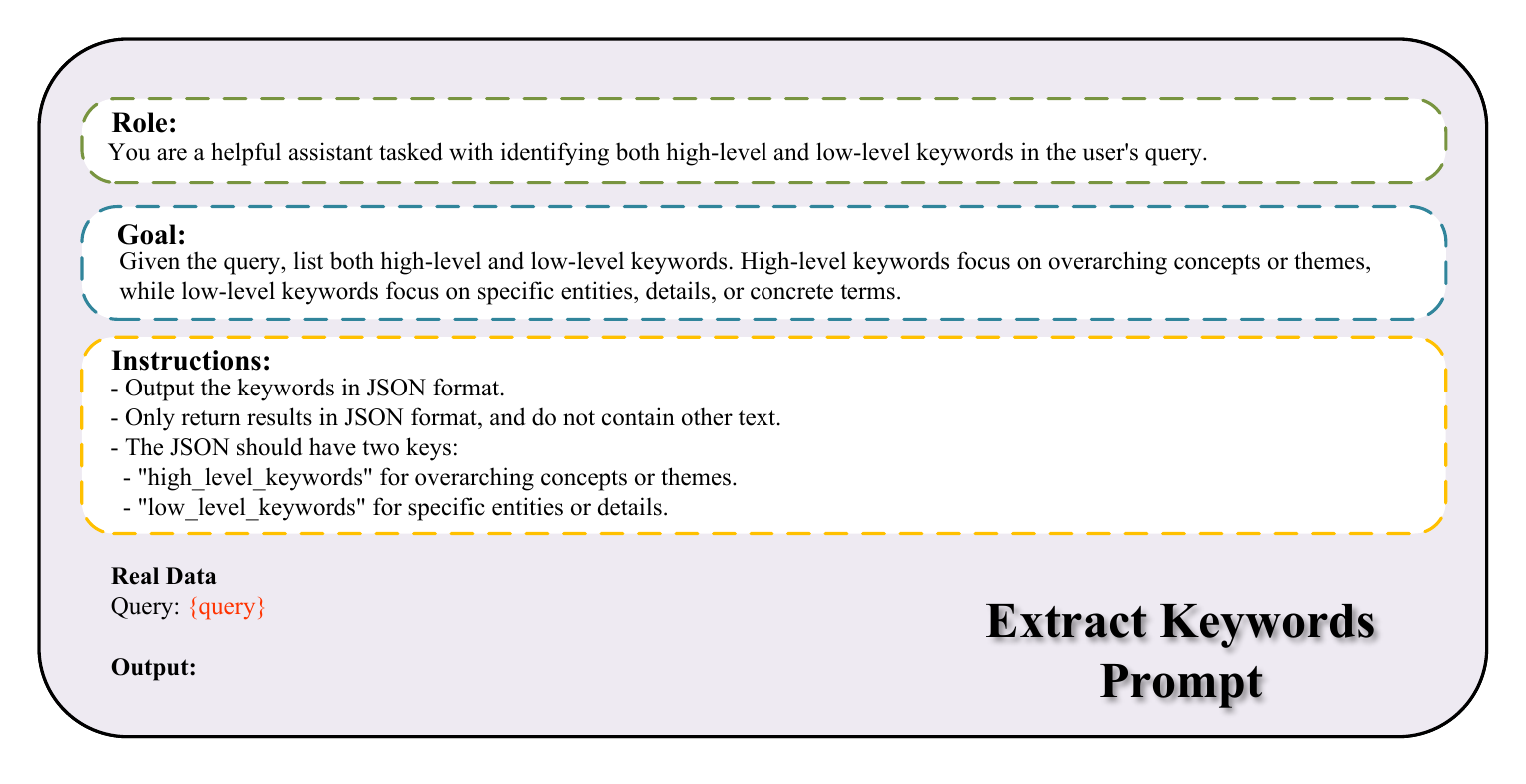}
\caption{Extract Key Prompt.}
\label{fig:Extract_key}
\end{figure}
The prompt template, as shown in Figure \ref{fig:Extract_key}, is used to extract both high-level and low-level keywords from the user's query. High-level keywords focus on overarching concepts or themes, while low-level keywords focus on specific entities, details, or concrete terms. The template requires returning the results in JSON format, containing two keys: ``high\_level\_keywords'' for overarching concepts or themes, and ``low\_level\_keywords'' for specific entities or details. It emphasizes returning only JSON results, without any additional text.

\subsection{Prompt for initial thought node}
\begin{figure}[htbp]
\centering
%\vspace{-1.2cm}
\includegraphics[width=\columnwidth]{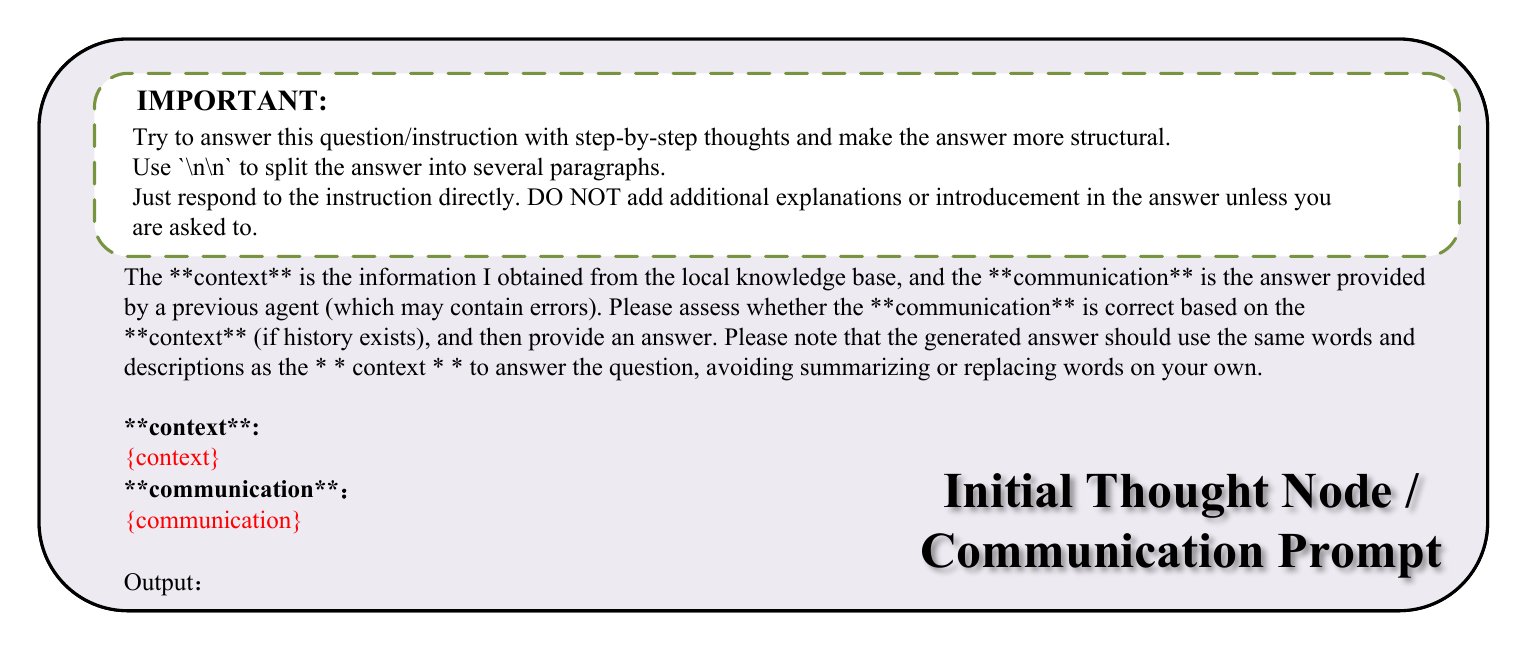}
\caption{Initial Thought Node.}
\label{fig:Initial_Thought_Node}
\end{figure}
The Initial Thought Node prompt template, as shown in Figure \ref{fig:Initial_Thought_Node}, is used to generate structured answers based on previously provided information and current instructions. The template requires the answer to be split into multiple paragraphs and directly respond to the given instructions, without adding extra explanations or introductions. It emphasizes using the previous context and communication information to ensure the structure and consistency of the answer. The output of the template will include context and communication content, following the direct response to the instructions.

\subsection{Prompt for Summary Node}
\begin{figure}[htbp]
\centering
%\vspace{-1.2cm}
\includegraphics[width=\columnwidth]{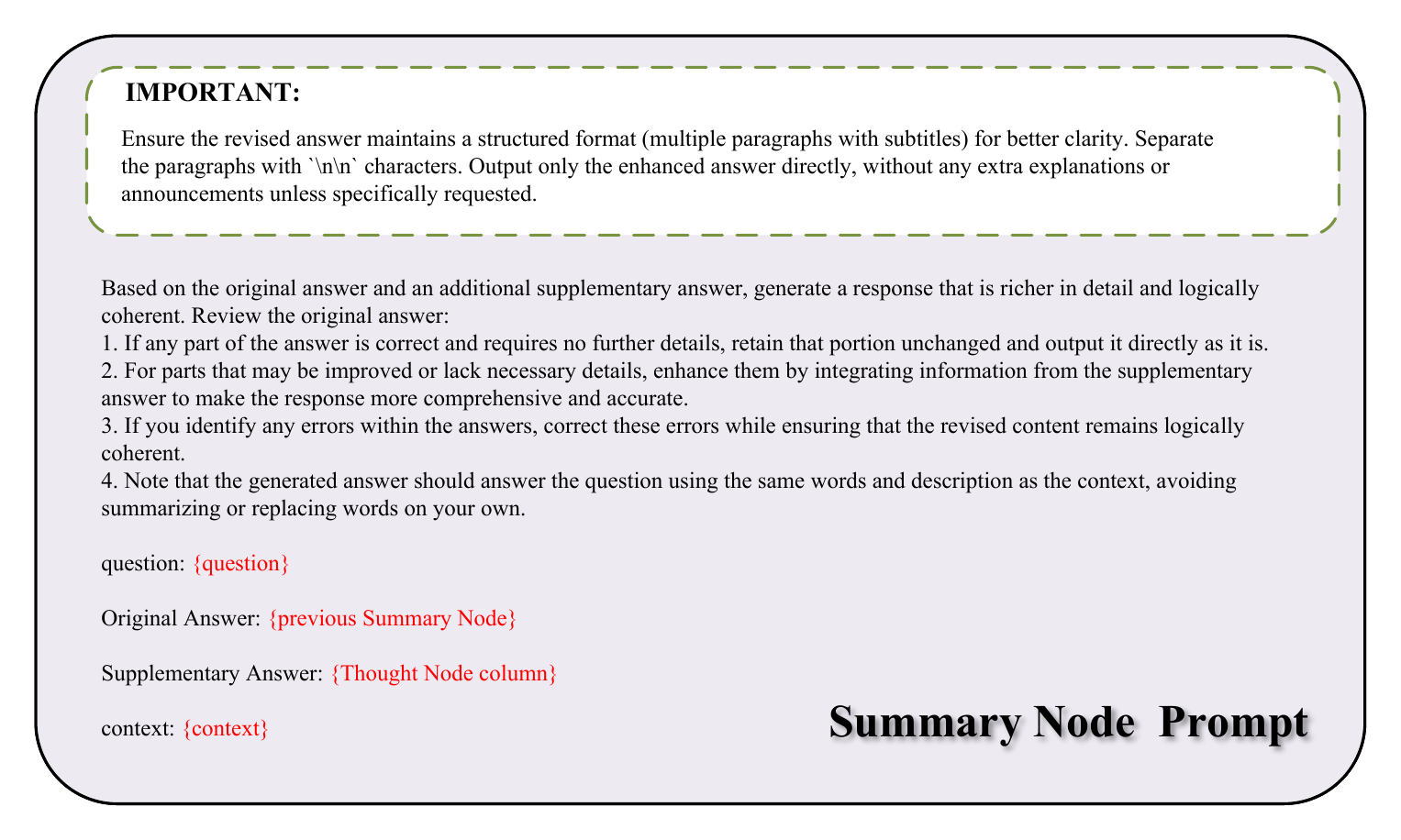}
\caption{Initial Thought Node.}
\label{fig:Summary_Node}
\end{figure}
The Summary Node prompt template, as shown in Figure \ref{fig:Summary_Node}, is used to generate more detailed and logically coherent answers based on the original answer and an additional supplementary answer. The template requires maintaining a structured format with multiple paragraphs. First, review the original answer, and if any part is correct and does not need further details, retain it unchanged. For parts that need improvement, enhance them by incorporating information from the supplementary answer to make the response more comprehensive and accurate. If any errors are found, they should be corrected while ensuring that the revised content remains logically coherent. Finally, the generated answer should use language consistent with the context and avoid summarizing or replacing words on your own.

\end{document}